\documentclass[11pt]{scrartcl}       
\usepackage[bottom=2.5cm,top=2.5cm,left=2.5cm,right=2.5cm]{geometry}

\usepackage{amsmath,amssymb,amsthm,graphicx,caption,cases}	
\usepackage[bottom]{footmisc}	
\usepackage{algorithmic}

\usepackage{abstract,amsfonts,amsbsy,amssymb,amstext,mathrsfs,latexsym,color,url,amsthm,xcolor,authblk,fontenc,bm}

\definecolor{NUSBlue}{RGB}{0,61,124} 
\definecolor{NUSOrange}{RGB}{239,124,0}
\usepackage[colorlinks=false,allbordercolors=NUSOrange]{hyperref}

\DeclareOldFontCommand{\bf}{\normalfont\bfseries}{\mathbf}

\definecolor{NUSBlue}{RGB}{0,61,124}   

\usepackage{tikz}
\usepackage{mathtools}

\def\nudge{.5}

\tikzset{axis/.style={ultra thin, Grey, -latex, shorten <=-\nudge cm, shorten >=-2*\nudge cm}}
\tikzset{line/.style={thick}}

\usepackage{textcomp}
\DeclareMathAlphabet\mathbfcal{OMS}{cmsy}{b}{n}
\usepackage{hyperref}
\usepackage[linesnumbered,vlined,ruled]{algorithm2e}
\usepackage{amsmath,amsfonts,amssymb,amsthm}
\theoremstyle{plain}
\newtheorem{thm}{Theorem}

\newtheoremstyle{cited}%
  {3pt}
  {3pt}
{\itshape}
  {}
  {\bfseries}
  {.}
  {.5em}
  {\thmname{#1} \thmnumber{#2} \thmnote{\normalfont#3}}
\theoremstyle{cited}
\newtheorem{citedthm}[thm]{Theorem}
\newtheorem{citedlem}[thm]{Lemma}
\newtheorem{citedcor}[thm]{Corollary}

\usepackage{graphicx,epstopdf,epsf,subfigure} 
\usepackage{cite}
\usepackage{tabularx}
\usepackage{array}
\usepackage{booktabs}
\usepackage{comment}

\begin{document}
\renewcommand*{\Authsep}{, }
\renewcommand*{\Authand}{, }
\renewcommand*{\Authands}{, }
\renewcommand*{\Affilfont}{\normalsize}   
\setlength{\affilsep}{2em}   
\title{Convergence rates of stochastic gradient method with independent sequences of step-size and momentum weight}
\date{\today}
\author{Wen-Liang Hwang}
\maketitle
\begin{abstract}
In large-scale learning algorithms, the momentum term is usually included in the stochastic sub-gradient method to improve the learning speed because it can navigate ravines efficiently to reach a local minimum. However, step-size and momentum weight hyper-parameters must be appropriately tuned to optimize convergence. 
We thus analyze the convergence rate using stochastic programming with Polyak's acceleration of two commonly used step-size learning rates: ``diminishing-to-zero" and ``constant-and-drop" (where the sequence is divided into stages and a constant step-size is applied at each stage) under strongly convex functions over a compact convex set with bounded sub-gradients. For the former, we show that the convergence rate can be written as a product of exponential in step-size and polynomial in momentum weight. Our analysis justifies the convergence of using the default momentum weight setting and the diminishing-to-zero step-size sequence in large-scale machine learning software. For the latter, we present the condition for the momentum weight sequence to converge at each stage.
\end{abstract}

\section{Introduction}

Optimization methods for large-scale machine learning involve the numerical computations of parameters for a system based on a sequence of mini-batch data. The stochastic programming (SG) method is popular for large-scale learning problems, and stochastic programming with momentum (SGM) is a state-of-the-art optimizer for benchmarks such as the image classification dataset ImageNet. The hyper-parameters of SGM optimizers (i.e., step-size and momentum sequences) require fine-tuning. This process can be time-consuming to obtain the values of the hyper-parameters using a grid search. Therefore, we aim to reduce the costs of tuning the hyper-parameters using convergence analysis on stochastic programming.

The gradient method solves 
\begin{align} \label{sto:problem}
\min_{\theta \in D} f(\theta)
\end{align}
via successively updating $\theta_j$ with $s_j \in \partial f(\theta_j)$ in closed convex set $D$, with initial guess $\theta_0 \in D$, using 
\begin{align*}
\theta_{j+1} = P_D(\theta_j - t_j s_j),
\end{align*}
where $t_j > 0$ is the step-size, and $P_D$ is non-expansive orthogonal projection operation onto $D$. 
A natural extension to stochastic programming is 
the minimization of the expected risk over the random variable $z$ with
\begin{align} \label{sto:emp}
\min_{\theta \in D} f(\theta) = E_{z} \{f(\theta;z)\}.
\end{align}
For example, the empirical risk minimization considers minimization of $f(\theta) = \frac{1}{N} \sum_{i=1}^N f(\theta; \{(x_i, y_i)\})$ based on the available data set $\{(x_i, y_i)\}$. If we draw $i \in [1, N]$ uniformly at random, then $f(\theta) = E_i\{f(\theta;  \{(x_i, y_i)\})\}$.  The SG method updates parameter $\theta_j$ with
\begin{align} \label{sto:stoupdate}
\theta_{j+1} = P_D(\theta_j - t_j g(\theta_j; z_j)),
\end{align}
where $z_j$ is a mini-batch sampled randomly from $\{(x_i, y_i)\}$, and $g(\theta_j; z_j)$ is an unbiased estimator of $s_j \in \partial f(\theta_j)$ in the mini-batch drawn under random variable $z_j$.
Motivated that the gradient method cannot navigate ravines efficiently to reach a local minimum, Polyak's momentum was proposed based on the heavy-ball method \cite{polyak1964some} for deterministic optimization. The global convergence and global bounds of the convergence rate of the method are established for convex optimization in \cite{ghadimi2015global}. The method updates the parameter $\theta_j$ with
\begin{align} \label{momentumupdate}
\theta_{j+1} = P_D( \theta_j - t_j s_j + \eta_j (\theta_j - \theta_{j-1}))
\end{align}
where $s_j \in \partial f(\theta_j)$, $t_j$ is the step-size, and $\eta_j < 1$ is referred to as the momentum weight.

Using momentum in stochastic gradient methods has become widespread in large-scale machine learning \cite{krizhevsky2012imagenet,sutskever2013importance,sun2019optimization}. Variants of momentum have been used in machine learning and demonstrated empirical successes in various tasks. In this paper, we focus on the SGM, an extension of Eq. (\ref{momentumupdate}) to stochastic programming that solves Eq. (\ref{sto:problem}) 
via successively updating $\theta_j$, with initial guesses $\theta_{-1}$ and $\theta_{0} \in D$, using 
\begin{align}\label{sto:mom1}
\theta_{j+1} = P_D(\theta_j - t_j g(\theta_j; z_j) + \eta_j (\theta_j - \theta_{j-1})).
\end{align}

In the context of SG and SGM, we focus on the sequence of momentum weights in Eq. (\ref{momentumupdate}) and Eq. (\ref{sto:mom1}) to convergence under two common step-size sequences (i.e., $t_j$): diminishing-to-zero with $\sum_j t_j = \infty$ and $\sum_j t_j < \infty$, and constant-and-drop learning rate \cite{bottou2018optimization} where the learning comprises several stages. In the pragmatic use of the latter, a stage applies a constant step size for several iterations and applies a factor drops this step size to refine the learning rate.

\subsection{Assumptions and contributions}

Throughout the paper, $\| x \|$ denotes the 2-norm of vector $x$, and $\theta^*$ denote the minimum of $m$-strongly convex function $f$. 

\noindent{\bf{Assumptions}}
Our analysis of convergence of SGM using various hyper-parameter sequences is established based on the following assumptions: 

\noindent{Assumption 1.} $D$ is a compact convex set in $\mathcal R^n$ with the diameter bounded by $L$ (i.e., $L \geq \sup_{\theta_i, \theta_j \in D} \{ \| \theta_i - \theta_j \|\}$). Function $f:\mathbb R^n \rightarrow \mathbb R$ is $m$-strongly convex over $D$. That is, for all $\theta_{i}$, $\theta_j \in D$ and $s_j \in \partial f(\theta_j)$, 
\begin{align} \label{mstrongly}
f(\theta_{i}) \geq f(\theta_j) + s_j^\top (\theta_{i} - \theta_j) + \frac{m}{2} \|\theta_{i} - \theta_j \|^2.
\end{align}
Let $\theta_i = \theta^*$ be the minimizer of $f$ in the interior of $D$. This implies that
\begin{align} \label{convexbound}
s_j^\top (\theta_j - \theta^*) \geq \frac{m}{2} \| \theta_j - \theta^* \|^2.
\end{align}
The assumption of strongly convex functions contrasts many optimization methods for DNN learning. This assumption, as stated in \cite{bottou2018optimization}, is based on the fact that most of the learning problem is regularized by a strongly convex function in the neighborhood convex subset $D$ of local minimizers. As indicated in \cite{zeng2019global,tung2024}, the solution to learn a non-convex objective function of a DNN comprises a sequence of the composition of sub-problems, each involving a strongly convex objective function.
Most previous analysis studies the number $j$ of iterations to sub-optimality; that is, $E\{f(\theta_j) - f^*\} \leq \epsilon$, where $f^*$ is the optimal value. By this assumption, we can consider the convergence of the mean-squared error sequence $E\{\|\theta_j - \theta^*\|^2\}$, because $E\{f(\theta_j) - f^*\} \rightarrow 0$ is equivalent to $E\{\|\theta_j - \theta^*\|^2\} \rightarrow 0$ for strongly convex functions. \\

\noindent{Assumption 2.} 
We can express $g(\theta_j; z_j) = s_j + n_j$ where 
$s_j$ and $n_j$ are random variables with $E\{n_j\} = 0$, as $g(\theta_j; z_j)$ is an unbiased estimator of $s_j \in \partial f(\theta_j)$ in the mini-batch drawn under random variable $z_j$. The variance of $n_j$ is bounded by $\sigma^2$, and $n_j$ is independent of $\theta_j$, $s_j$, and $n_i$, with $i \neq j$. This assumption of an unbiased estimate of a sub-gradient/gradient of $f$ has been commonly applied to stochastic approximation using the stochastic sub-gradient/gradient method \cite{strassen1965existence,hazan2014beyond,bottou2018optimization}. \\

\noindent{Assumption 3.} 
For any $\theta$ in $\mathcal D$ and any $s \in \partial f(\theta)$, we assume $\|s \| \leq \sqrt M$.
This assumption implies  
 \begin{align} \label{gradientbound}
 |f(\theta_1) - f(\theta_2) |\leq \sqrt{M} \|\theta_1 - \theta_2\| \text{ for any $\theta_1, \theta_2 \in \mathcal D$}.
 \end{align}
 
\noindent Note that Assumptions 2 and 3 are also adopted in stochastic optimization of non-smooth functions \cite{nemirovski2009robust,davis2019stochastic,mai2020convergence}. \\

\noindent{\bf{Contributions}} 

We show the worst-case analysis of the convergence of the mean-squared errors for various momentum weight sequences under the diminishing-to-zero step-size sequence with $\sum_j t_j = \infty$ and $\sum_j t_j < \infty$. Our results write the convergence rates as functions of independent hyper-parameter sequences: step-size $\{t_j\}$ and momentum weight $\{\eta_j\}$. Decoupling the step-size and momentum weight sequences to convergence provides a simple way to explain and fine-tune the parameters in learning tasks. \\

\noindent{$\bullet$} Specifically for diminishing-to-zero step-size sequence, the convergence rate for mean-squared-errors is 
\begin{align*}
{\cal O}(e^{-(m \sum_i^{N-1} t_i + \frac{m^2}{2} \sum_j t_j^2) } (1  +\sum_{i=1}^{N}\eta_i)).
\end{align*}
In the above, the first term, ${\cal O}(e^{-(m \sum_i^{N-1} t_i + \frac{m^2}{2} \sum_j t_j^2) })$, characterizes the convergent rate for SG. This result indicates that under the worst-case analysis, the best convergence rate order of an SGM is the same as that of SG (i.e., when $\sum_i \eta_i < \infty$). However, this result should not be against the use of the momentum. Using momentum in practice for DNN learning converges fast and can avoid trapping in a bad local minimum. Our analysis supports the default setting $\eta_j = k t_j$ in Adam \cite{kingma2014adam}, Tensorflow \cite{abadi2016tensorflow}, and PyTorch \cite{paszke2017automatic} software, and explains improving performance when training DNNs \cite{chen2019demon} and generative adversarial networks (GANs) \cite{arjovsky2017wasserstein}
by decreasing momentum weights. \\

\noindent{$\bullet$} For constant-and-drop learning rate (also referred to as multi-stage SGM learning) that uses the suffix average strategy at each stage, our analysis shows that the mean-squared-error sequence can converge, as long as momentum weights $\eta_j$ at the stage satisfy $\frac{\sum_{i=0}^j \eta_i}{(j+1)} \rightarrow 0$ with $\eta_i < 1$. For this case, the momentum weight can be ${\cal O}(\frac{1}{(j+1)^\beta})$ with $0 < \beta \leq 1$, where $j$ is the number of updates at a stage. This result justifies the practical use that keeps momentum weights unchanged or gradually decreasing at each stage. \\

This paper organizes the remainder as follows. Section \ref{review} reviews some related works. Section \ref{analysis} presents analytical results. Section \ref{conclusion} offers concluding remarks and addresses relevant issues.

\section{Related works} \label{review}

The convergence rate of SGM and SG varies under various assumptions about objective functions, step-size selection, and iterator/estimator derivability. 


We begin by reviewing some results using SG. 
Matching the lower bound analysis presented in \cite{nemirovskij1983problem,agarwal2009information}, the optimal algorithm in \cite{hazan2014beyond} attains convergence rate of $E\{f(\theta_N) - f(\theta^*)\}$ for strongly convex functions with bounded gradients, with ${\cal O}(\frac{1}{N})$ approximation solution after $N$ gradient updates. Using SG of step-sizes $\frac{\gamma}{j+1}$,\cite{rakhlin2011making,YuxinChen2018} obtain this convergence rate for smooth, strongly convex functions with bounded gradients.
For non-smooth, strongly convex functions with bounded sub-gradients, \cite{shamir2013stochastic} using SG with step-sizes $\frac{\gamma}{j+1}$ obtains $E\{\|\bar{\theta}_N- \theta^* \|\} = {\cal O}(\frac{\log N}{N})$, where $\bar{\theta}_N$ is the weighted average with a weight of $j+1$ for iterate $\theta_j$, while  \cite{lacoste2012simpler} improves this bound to ${\cal O}(\frac{1}{N})$.
\cite{shalev2011pegasos} applies SG to solve the optimization problem with strongly convex and non-smooth objective functions cast by the SVM. 
When convex functions are Lipschitz continuous gradients, satisfy the Polyak-Lojasiewicz condition (i.e., $2 \mu (f(\theta) - f(\theta^*))  \leq \|\nabla f(\theta) \|^2$), \cite{yuan2019stagewise} demonstrates ${\cal O}(\frac{1}{N})$ convergence rate using SG with step-size sequence $\frac{2t+1}{2\mu (t+1)^2}$.

The momentum method is a technique for accelerating the gradient method that accumulates a velocity vector in directions of persistent reduction in the objective across iterations. 
Besides Polyak's momentum, Nesterov's \cite{nesterov2003introductory} has also been studied under stochastic programming for training DNNs \cite{sutskever2013importance}. When using Polyak's momentum, a ball rolling down a hill follows exclusively the slope information. In contrast, under  Nesterov's momentum, the ball slows down before the hill slopes up again using an extrapolation of the ball's location \cite{ruder2016overview}. 
Due to the empirical success of employing momentum in large-scale DNN learnings, the variants of momentum grow (e.g., synthesized Nesterov variants (SNV) \cite{lessard2016analysis}, triple momentum \cite{van2017fastest}, accelerated SGD (AccSGD) \cite{kidambi2018insufficiency}, and quasi-hyperbolic momentum (QHM) \cite{ma2018quasi}). Variants of assumptions regarding the objective functions (e.g., strongly convex functions; smooth, non-convex functions with Lipschitz continuous gradients, non-smooth convex functions; bounded sub-gradients; and weakly convex functions, or a combination of some of them) have been adopted in the analysis of the convergence in terms of the number of iterations to sub-optimality or a stationary/critical point.

Polyak's acceleration can be seen as a discretization of ordinary differential equations (ODEs) and studied using a continuous dynamic model where momentum weight was analogous to the mass of Newtonian particles that move through a dense medium in a conservative force field \cite{qian1999momentum}. 
Using a discretization of the heavy ball ODE with additional perturbation, 
the exponential convergence for objective functions under Polyak-Lojasiewicz inequality can be established \cite{gess2023convergence}. Using the ODE proposed in \cite{da1810general} that unifies Polyak's and Nesterov's momentum methods, \cite{barakat2021stochastic} establishes the stability and almost sure convergence of the iterates to the set of critical points for smooth and non-convex functions.

Our iterator update using Eq. (\ref{momentumupdate}) assigning independent step-size and momentum weight sequences stands in contrast with the normalized SGM with the following update:
\begin{align} \label{Watao}
\begin{cases}
z_j = \beta_j s_j + (1-\beta_j) z_{j-1} \\
\theta_{j+1} = P_D(\theta_j - \alpha_j z_j).
\end{cases}
\end{align}
This update relates step-size and momentum weight sequences
with $t_j = \alpha_j \beta_j$ and $\eta_j = \alpha_j (1- \beta_j)$; hence, $\eta_j = \frac{1-\beta_j}{\beta_j} t_j$. 
A general formulation of QHM introducing one more parameter $v_j$ than the SGM is proposed in \cite{gitman2019understanding} to give a unified treatment of SG,  SGM, and several momentums, under the assumptions of smooth functions with Lipschitz continuous gradients and bounded gradients, in which $v_j \in [0, 1], \beta_j \in [0, 1), \alpha_j > 0$, and 
\begin{align}
\begin{cases}
m_j = (1-\beta_j)  \nabla f(\theta_j)  + \beta_j m_{j-1} \\
\theta_{j+1} = \theta_j - \alpha_j [(1-v_j) \nabla f(\theta_j) + v_j m_j].
\end{cases}
\end{align}
The parameter $v_j$ interpolates between SG ($v_j= 0$) and normalized SGM ($v_j = 1$). Setting $v_j=1$, $\beta_j = \eta_j$ and $\alpha_i (1-\beta_j) = t_j$ yields SGM. Using analysis on QHM, \cite{liu2020improved} shows that the normalized SGM (Eq. (\ref{Watao})) achieves the same convergence bounds as SG \cite{liu2020improved}. Under the assumptions of weakly-convex functions on closed convex sets and bounded sub-gradients, \cite{mai2020convergence} shows that the convergence rate to a stationary point is in order of ${\cal O}(\frac{1}{\sqrt{1 + K}})$, where $K$ is the number of iterations (equivalently, order of ${\cal O}(\frac{1}{\epsilon^2})$ to sub-optimality) by adopting the technique in \cite{davis2019stochastic} for convergence of stochastic methods on weakly convex problems.

The constant-and-drop learning rate is one of the most widely adopted strategies in large-scale machine learning. The learning involves several stages. At each stage, the strategy requires a fixed step-size, a sequence of momentum weights, stage length (i.e., the number of iterations), and the method to derive the estimator (e.g., last update and suffix average). After running a stage to proceed to the next stage, the pragmatic usage doubles the number of iterations running for a fixed step-size after reducing the step-size to half its previous value. At the same time, the momentum weights are re-initialized. The momentum weights run almost constant or slowly decreasing \cite{ghadimi2013optimal} in a stage. 
For strongly convex functions with Lipschitz continuous gradients and limits on first and second moments, \cite{bottou2018optimization} shows the convergence of SG at a stage as follows:
\begin{align}  \label{piecewise0}
\lim_{j \rightarrow \infty} E\{|f(\theta_j) -f(\theta^*)|^2\} \leq  ka,
\end{align}
where $k$ is a system-related parameter, $a$ is the constant step-size. Under the assumption of Lipschitz continuous gradients, \cite{liu2020improved} demonstrates the convergence of a multi-stage algorithm by allowing large step sizes in the first few stages and smaller step sizes in the final stages. The multi-stage algorithm in \cite{aybat2019universally} proceeds in stages using a stochastic version of Nesterov’s accelerated method with geometrically growing lengths and shrinking step sizes for the following stages. Given a specific choice of length at the first stage and the computational budget, that algorithm can achieve the optimal error bounds in the deterministic and stochastic case for smooth, strongly convex functions with noisy gradient estimates.

\section{Convergence analysis} \label{analysis}

We consider SGM with sequences of step-size $\{t_j | \sum_j t_j = \infty \text{ and } \sum_j t_j^2 < \infty\}$ and momentum weight $\{\eta_j<1\}$ that lead to  convergence in the mean-square sense. 
We first characterize the convergence rate of SG. 
\begin{citedlem} \label{SGcon}
Suppose Assumptions 1-3 hold. Let $m$ be defined according to Eq. (\ref{mstrongly}). Using diminishing-to-zero step-size sequence with $\sum_i t_i = \infty$ and $\sum_i t_i^2 < \infty$, the mean-squared error derived by SG using Eq. (\ref{sto:stoupdate}) converges to zero. Precisely, there exist $j_0$ and $c_0$ (depending on $j_0$) such that for all $N \geq j_0$, the converge rate satisfies
\begin{align} \label{SGupper}
E\{\|\theta_{N} -\theta^*\|^2\} \leq c_0  e^{ -\sum_{j=1}^{N} (t_j+ 2 t_j^2)}.
\end{align}
If letting step-size sequence be $t_j = \frac{\gamma}{(j+1)^{\alpha}}$ with $\gamma = \frac{1}{m}$ and $0 < \alpha \leq 1$, then there exists $j_1(\alpha)$ and constant $c_1(\alpha)$ (depending on $j_1(\alpha)$) such that for $N \geq j_1(\alpha)$, the convergence rate is
\begin{align} \label{SGupper}
E\{\|\theta_{N} -\theta^*\|^2\} \leq \frac{c_1(\alpha)}{N+1}.
\end{align}
\end{citedlem}

\proof

See Appendix~\ref{SGconvergence}.

\qed

This result achieves the lower bound of  \cite{nemirovskij1983problem,agarwal2009information}. Informally, the lower bound indicates that no algorithm can perform $t$ queries to noisy first-order oracles to achieve a result better than the order of $\frac{1}{t}$ when minimizing strongly convex functions \cite{YuxinChen2018}. For SGM, we need the following lemma.

\begin{citedlem}\label{sto:rmsthm}
Suppose Assumptions 1-3 hold. Let $t_i m \in (0, 1]$ for all $i$ where $m$ is defined in Eq. (\ref{mstrongly}). The mean-squared error derived by SGM using Eq. (\ref{momentumupdate}) is upper bounded, with
\begin{align} \label{sto:expand34}
E\{ \| \theta_{N} - \theta^*\|^2\} & \leq \left( \prod_{i=2}^{N-1}  (1-t_im) \right)E\{ \|\theta_2 - \theta^*\|^2\} + t_{N-1}^2 (M + \sigma^2)  \nonumber  \\
& + \sum_{i=2}^{N-2} \left( \prod_{k=i+1}^{N-1}(1-t_km) \right) t_i^2 (M + \sigma)^2 \nonumber \\ 
& + \sum_{i=2}^{N-2} \left( \prod_{k=i+1}^{N-1}(1-t_km) \right) 2 \eta_i E\{ (\theta_i - \theta^* - t_i s_i)^\top (\theta_i - \theta_{i-1})\} \nonumber \\ 
& + \sum_{i=2}^{N-2} \left( \prod_{k=i+1}^{N-1}(1-t_km) \right) \eta_i^2 E\{\| \theta_i - \theta_{i-1} \|^2 \} \nonumber \\ 
&  + 2 \eta_{N-1} E\{ (\theta_{N-1} - \theta^* - t_{N-1} s_{N-1} )^\top (\theta_{N-1}- \theta_{N-2}) \} \nonumber \\ 
& + \eta_{N-1}^2 E \{ \| \theta_{N-1}- \theta_{N-2}\|^2 \}.
\end{align}

\end{citedlem}
\proof
Refer to Appendix \ref{SGM-stepsize1}.

\qed

In the following subsections, we discuss two step-size sequences for SGM.

\subsection{Diminishing-to-zero step-sizes}

We demonstrate the condition on momentum weight sequences to achieve the convergence of SGM using diminishing-to-zero step-size sequence with $\sum_j t_j = \infty$ and $\sum_j t_j^2 < \infty$.

\begin{citedthm} \label{dthm}
Under Assumptions 1-3, $t_i m \in (0, 1]$ for all $i$, $\sum_i t_i = \infty$, $\sum_i t_i^2 < \infty$, and $\eta_i < 1$. Let $\eta_i$ be a non-increasing diminishing-to-zero sequence. Then, there exists $j_0$ and $c_0$ (depending on $j_0$) such that for all $N \geq j_0$, the convergence rate of the mean-squared error of SGM using Eq. (\ref{momentumupdate}) is upper bounded with 
\begin{align} \label{diminishingt}
E\{\|\theta_{N} -\theta^*\|^2\} \leq c_0 e^{-(m \sum_i^{N-1} t_i + \frac{m^2}{2} \sum_j^{N-1} t_j^2) } (1  +\sum_{i=1}^{N}\eta_i).
\end{align}
Therefore, the condition on step-size and momentum weight sequences for SGM to converge to zero is 
$e^{-(m \sum_i^{N-1} t_i + \frac{m^2}{2} \sum_j^{N-1} t_j^2)} \sum_i^N \eta_i \rightarrow 0$ as $N \rightarrow \infty$.
Moreover, setting $t_j = \frac{\gamma}{(j+1)^{\alpha}}$ with $\gamma = \frac{1}{m}$ and $0 < \alpha \leq 1$, we can obtain the following converge rate of $E\{\|\theta_{N} -\theta^*\|^2\}$:  \\
(i) ${\cal O}(\frac{1}{N+1})$ if $\sum_j^N \eta_j < \infty$;  \\
(ii) ${\cal O}(\frac{\log (N+1)}{N+1})$ if $\sum_j^N \eta_j = \infty$ and $\eta_j = \frac{1}{j+1}$; \\
(iii) ${\cal O}(\frac{1}{(1-\beta) (N+1)^\beta})$ if $\sum_j^N \eta_j = \infty$ and $\eta_j = \frac{1}{(j+1)^\beta}$ for $0 < \beta < 1$.
\end{citedthm}
\proof

Refer to Appendix \ref{SGMconvergence}.

\qed

This analysis clarifies the relationship between SGM's convergence rate and the decay rate of momentum weights under the diminishing-to-zero step-size sequence. For fast decay momentum weight sequence, with $\sum_j \eta_j \rightarrow \infty$, the convergence rate of SGM is in the same order as SG, whereas for slow decay sequence, with ${\cal O}(\frac{1}{(j+1)^\beta}$ with $0 < \beta \leq 1$, the convergence rate of SGM is inferior to SG.

\subsection{Constant-and-drop step-sizes}

In practice, the constant-and-drop learning strategy is often employed by running the SG and SGM with a fixed step size in the update until progress stalls; then, a smaller step-size is selected, and the process is repeated. 
This strategy divides the learning sequence into stages and applies a constant step-size at each stage.

Suppose $N$ updates, $\theta_1, \cdots, \theta_N$, is obtained at a stage. The strategy of the last update uses $\theta_N$ as the estimator of $\theta^*$ at the stage, whereas the strategy of the suffix average update uses $\frac{1}{N} \sum_i \theta_i$ as the estimator of $\theta^*$. 
Considering SG using a fixed step-size $\frac{1}{m} > a > 0$ at a stage, we derive in Appendix  \ref{appendixC} that there exists finite but sufficiently large $N_a$, depending on $a$, such that for all $N \geq N_a$, the mean-squared-error of SG using Eq. (\ref{momentumupdate}) and the last update is upper bounded with
\begin{align} \label{constantM}
E\{\|\theta_N -\theta^*\|^2\}  \leq  \frac{2(M + \sigma^2)}{m} a.
\end{align}
This result is consistent with Eq. (\ref{piecewise0}), which assumes Lipschitz continuous gradients. 
The suffix average update can extend Eq. (\ref{constantM}) to SGM. 

\begin{citedthm} \label{piecewisemomlem}
Under Assumptions 1-3 and the constant-and-drop learning strategy. Let momentum weight sequence in a stage of constant step-size $a$ with $\frac{1}{m} > a > 0$ satisfy $\eta_j < 1$ and $\frac{\sum_{i=0}^j \eta_i}{(j+1)} \rightarrow 0$ as $j$ approaches infinite. Then, there exists $N_a$, and for $N \geq N_a$,  the mean-squared-error in the stage derived by SGM using Eq. (\ref{momentumupdate}) is upper bounded with 
\begin{align} \label{constantandstop1}
E\{\|\frac{1}{N+1} \sum_{i=0}^{N} \theta_i -\theta^*\|^2\}  \leq  \frac{2(M + \sigma^2)}{m} a.
\end{align}
\end{citedthm}
\proof
See Appendix \ref{piecewisemom}.

\qed

This theorem indicates that mean-squared error can converge if  $\sum_i^j \eta_i$ is of order $o(j+1)$, which means that $\sum_i \eta_i$ goes to zero faster than $j+1$ as $j$ approaches infinity. Accordingly, it justifies the practical selection of slowly decaying momentum weight sequence in a stage, due to the fact that $\eta_i = \frac{c}{(1 + j)^\beta}$ with $0 < \beta \leq 1$ is of order $o(j+1)$, which can be derived using Eq. (\ref{slowhar}) for $0 < \beta < 1$, and using L'Hopital rule to obtain $\frac{\log x}{x} \rightarrow 0$ as $x$ approaches infinite for $\beta = 1$. The following result is obvious.

\begin{citedcor} \label{thmconstanddrop}
Suppose the assumption for Theorem \ref{piecewisemomlem} holds. Let $\{a_i\}$ be a strictly decreasing sequence to zero, and $a_j$ is the constant step-size at stage $j$. Let's apply the suffix average update to the constant-and-stop strategy. Denote $\bar \theta_j$ as the average at stage $j$ (the average of $N$ updates with $N \geq N_{a_j}$), then
\begin{align} \label{SGMconstantanddrop2}
\lim_{j \rightarrow \infty} E\{\|\bar \theta_{j} -\theta^*\|^2\}  \rightarrow 0.
\end{align}
\end{citedcor}

\section{Conclusions and discussions} \label{conclusion}

This study resolves issues associated with SGM with parameter tuning step-sizes and momentum weights to achieve the mean-squared convergence to $\theta^*$ under strongly convex functions, bounded sub-gradients, and bounded $\|\theta_j - \theta^*\|$. We present the worst-case analysis of independent step-size and momentum weight sequences to the convergence of mean-squared errors to zero. Using the diminishing-to-zero step-size sequences with $\sum_i t_i = \infty$ and $\sum_i t_i^2 < \infty$,  
our analysis reveals that the convergence rate of SGM is the best as good as that of SG. So, what makes SGM popular pragmatically in large-scale machine learning might not be due to the worst-case behavior (if our assumptions are realistic). We conjecture that the objective functions in DNNs might contain plenty of ravines, which can trap SG often, but not SGM, or that the expected complexity (the expectation over the space of all possible inputs) of SGM is superior to that of SG, or the constant to the order of the worst-case analysis of SGM is smaller. These unresolved issues are worth further studying. Despite the diminishing-to-zero step-sizes, for  "constant-and-drop" learning rate strategy, we present condition $\frac{1}{N} \sum_j^N \eta_j \rightarrow 0$ on momentum weight sequence at a stage for the suffix average to convergence at the stage. We can connect this condition on $\eta_j$ to independently and identically distributed (i.i.d.) random variables with mean zero and finite variables via the central limit theory to provide a future direction for studies, in which parameters are sampled from i.i.d. random variables. \\

\noindent{\bf{Acknowledgement:}} 
Wen-Liang Hwang would like to thank Ming-Yu Chung, Cheng-Han Yeh, and Yu-Ting Huang for inspiring discussions on SGM's convergence rate.

\bibliographystyle{ieeetr}
\bibliography{PSGM}

\appendix

\section{Proof of Lemma \ref{SGcon} } \label{SGconvergence}

Put together (\ref{convexbound}) and  
\begin{align} \label{sto:maineq2}
\|\theta_{j+1} -\theta^*\|^2 &  =\|P_D(\theta_j  - t_j  s_j - t_j n_j) - P_D(\theta^*) \|^2  \nonumber\\
& \leq \| \theta_j   - \theta^* -t_j  s_j - t_j n_j \|^2
\end{align}
enables us to express 
Eq. (\ref{sto:maineq2}) as follows:
\begin{align} 
E\{\|\theta_{j+1} -\theta^*\|^2\} 
& \leq E\{\| \theta_j   - \theta^* -t_j  s_j - t_j n_j \|^2\} \nonumber \\
& \leq E\{\|\theta_{j} -\theta^*\|^2\}+ t_j^2 (M + \sigma^2) - 2 t_j E\{s_j^\top(\theta_j- \theta^*)\} \nonumber \\
& \leq E\{\|\theta_{j} -\theta^*\|^2\}+ t_j^2 (M+ \sigma^2) -  t_j m E\{\|\theta_j - \theta^*\|^2\} \nonumber \\
& = (1-t_j m) E\{\|\theta_{j} -\theta^*\|^2\})+ t_j^2 (M + \sigma^2).  \label{expect}
\end{align}
Rewriting Eq. (\ref{expect}) for $j=1$ to $N$ and assuming $t_j m \in (0, 1]$ for all $j$ yields
\begin{align} 
E\{\|\theta_{N} -\theta^*\|^2\} & \le (M + \sigma^2) \sum_{i=1}^{N-2} [t_i^2 \prod_{j=i+1}^{N-1}(1-t_j m)] + [\prod_{j=1}^{N-1} (1- t_j m)] E\{\|\theta_{1} -\theta^*\|^2\} \nonumber  \\
& + (M + \sigma^2) t_{N-1}^2  \label{SGkey1}  \\
& \leq \prod_{j=1}^{N-1} (1- t_j m)  [(M + \sigma^2) \sum_{i=1}^{N-2} t_i^2 + E\{\|\theta_{1} -\theta^*\|^2\}] + (M + \sigma^2) t_{N-1}^2 \nonumber  \\
&  \leq e^{ -(m \sum_{j=1}^{N-1} t_j + \frac{m^2}{2} \sum_{j=1}^{N-1}t_j^2)}[(M + \sigma^2) \sum_{i=1}^{N-2} t_i^2 +  E\{\|\theta_{1} -\theta^*\|^2\}]+ (M + \sigma^2) t_{N-1}^2. \label{keyeq1}
\end{align}
The last inequality uses the fact that $e^{-(x+\frac{x^2}{2})} \geq 1-x$ with $x \in [0, 1]$. Using the fact that $\sum_j t_j^2 < \infty$, $\sum_j t_j = \infty$, and $t_j \rightarrow 0$ as $j$ approaches infinity, we can obtain the order of mean-squared-error convergence ${\cal O}(e^{ -\sum_{j=1}^{N} (t_j+ 2 t_j^2)})$; i.e., there exist $j_0$ and constant $c_0$ (depending on $j_0$) such that, for all $N \geq j_0$,
\begin{align} \label{SGMrate0}
E\{\|\theta_{N} -\theta^*\|^2\} \leq c_0  e^{ -\sum_{j=1}^{N} (t_j+ \frac{t_j^2}{2})}.
\end{align}

\noindent i) $\alpha=1$: Substituting $t_j = \frac{\gamma}{j+1}$ with $\gamma = \frac{1}{m}$ into (\ref{SGMrate0}), we obtain the order of convergence ${\cal O}(\frac{1}{N+1})$; i.e., there exist $j_1$ and constant $c_1$ (depending on $j_1$), such that for $N \geq j_1$ 
\begin{align*} 
E\{\|\theta_{N} -\theta^*\|^2\} \leq c_0  e^{ -\sum_{j=1}^{N} (t_j+ \frac{t_j^2}{2})} \leq \frac{c_1}{N+1}.
\end{align*}
The last inequality is derived using the following two inequalities:
\begin{align} \label{SG1}
\sum_{j=1}^N \frac{1}{j+1} 
\geq \int_2^{N+1} \frac{1}{x} dx = \log (N+1) - \log 2
\end{align}
and 
\begin{align} \label{SG2}
\frac{1}{2} \sum_{j=1}^N \frac{1}{(j+1)^2} \geq \frac{1}{2}(\frac{1}{2} - \frac{1}{N+1}).
\end{align}
Combining Eqs. (\ref{SG1}) and (\ref{SG2}), we conclude that for all $N \geq N_0$, there exists $k$ (depending on $N_0$) such that
\begin{align*}
-\sum_{j=1}^N [\frac{1}{j+1} + \frac{1}{2 (j+1)^2}]  \leq \log \frac{k}{N+1};
\end{align*}
equivalently, 
\begin{align*}
e^{-\sum_{j=1}^N [\frac{1}{j+1} + \frac{1}{2 (j+1)^2}]} \leq \frac{k}{N+1}.
\end{align*}

\noindent ii) $0 < \alpha < 1$: 
Substituting $t_j = \frac{\gamma}{(j+1)^\alpha}$ with $\gamma = \frac{1}{m}$ into Eq. (\ref{SGMrate0}), we obtain the order of convergence ${\cal O}(\frac{1}{N+1})$; i.e., there exist $ j(\alpha)$ and constant $c(\alpha)$ (depending on $j(\alpha)$), such that for all  $N \geq j(\alpha)$: 
\begin{align*} 
E\{\|\theta_{N} -\theta^*\|^2\} \leq c_0  e^{ -\sum_{j=1}^{N} (t_j+ 2 t_j^2)}  \leq c_0  e^{ -\sum_{j=1}^{N} t_j} \leq \frac{c(\alpha)}{N+1}.
\end{align*}
The last inequality is derived in accordance 
\begin{align} \label{SG11}
\sum_{j=1}^N \frac{1}{(j+1)^{\alpha}} 
\geq \int_2^{N+1} \frac{1}{x^{\alpha}} dx =\frac{1}{1-\alpha}[(N+1)^{1-\alpha} - 2^{1-\alpha}].
\end{align}
This implies that 
\begin{align*}
\log (N+1)- \sum_{j=1}^N \frac{1}{(j+1)^{\alpha}}  & \leq \log (N+1) - [\frac{1}{1-\alpha}[(N+1)^{1-\alpha} - 2^{1-\alpha}] \\
& \leq \log (N+1) - [(N+1)^{1-\alpha} - 2^{1-\alpha}].
\end{align*}
It is clear that $\log (N+1) - [(N+1)^{1-\alpha} - 2^{1-\alpha}] \rightarrow -\infty$.
Hence, for all sufficiently large $N$ (depending on $\alpha$), the following holds: 
\begin{align*}
\log (N+1) \leq \sum_{j=1}^N \frac{1}{(j+1)^{\alpha}}.
\end{align*}
Equivalently,  
\begin{align*}
\frac{1}{N} \geq e^{-\sum_{j=1}^N \frac{1}{(j+1)^{\alpha}}}.
\end{align*}

\qed

\section{Proof of Lemma \ref{sto:rmsthm}} \label{SGM-stepsize1}

Using the non-expansive orthogonal projection operator to $D$, $P_D$, we can obtain 
\begin{align*} 
\|\theta_{j+1} - \theta^*\|^2 & = \|P_D(\theta_j + \eta_j (\theta_j - \theta_{j-1}) - t_j (s_j + n_j) - P_D(\theta^*)\|^2 \\
& \leq  \| \theta_j - \theta^* - t_j (s_j + n_j)  + \eta_j (\theta_j - \theta_{j-1}) \| ^2.
\end{align*}
We can apply the derivations similarly to (\ref{expect}) to yield 
\begin{align}  \label{expectM}
E\{\|\theta_{j+1} - \theta^*\|^2 \} & \leq (1 -t_j m) E\{\|\theta_j - \theta^*\|^2 \}+ t_j^2 (M+ \sigma^2)  \nonumber \\
&  + 2 \eta_j E\{(\theta_j -  \theta^* - t_j s_j )^\top (\theta_j - \theta_{j-1}) \}+ \eta_j^2 E\{\|\theta_j - \theta_{j-1}\|^2\}.
\end{align}

Rewriting Eq. (\ref{expectM}) for $j=2$ to $N-1$ and using the fact that $t_k m \in [0, 1]$ yields Eq. (\ref{sto:expand34}). \qed

\section{Proof of Theorem~\ref{dthm}} \label{SGMconvergence}

Eq. (\ref{diminishingt}): Because that $e^{-(x + \frac{x^2}{2})} \geq 1-x$ with $x \in [0, 1]$ and $t_i m \in (0, 1]$, Eq. (\ref{sto:expand34}) can be bounded:
\begin{align} \label{sto:expand}
E\{ \| \theta_{N} - \theta^*\|^2\} & \leq \left( \prod_{i=2}^{N-1}  (1-t_im) \right)E\{ \|\theta_2 - \theta^*\|^2\}  \nonumber  \\
& + \sum_{i=2}^{N-2} \left( \prod_{k=i+1}^{N-1}(1-t_km) \right) t_i^2 (M + \sigma)^2 \nonumber \\ 
& + \sum_{i=2}^{N-2} \left( \prod_{k=i+1}^{N-1}(1-t_km) \right) 2 \eta_i E\{ (\theta_i - \theta^* - t_i s_i)^\top (\theta_i - \theta_{i-1})\} \nonumber \\ 
& + \sum_{i=2}^{N-2} \left( \prod_{k=i+1}^{N-1}(1-t_km) \right) \eta_i^2 E\{\| \theta_i - \theta_{i-1} \|^2 \} \nonumber \\ 
&  + 2 \eta_{N-1} E\{ (\theta_{N-1} - \theta^* - t_{N-1} s_{N-1} )^\top (\theta_{N-1}- \theta_{N-2}) \}  + t_{N-1}^2 (M + \sigma^2) \nonumber \\ 
& + \eta_{N-1}^2 E \{ \| \theta_{N-1}- \theta_{N-2}\|^2 \} \nonumber \\
& \leq \left[ \prod_{i=2}^{N-1}  (1-t_im) \right] [E\{ \|\theta_2 - \theta^*\|^2\}+\sum_{i=2}^{N-2} t_i^2 (M + \sigma)^2 \nonumber \\
& +\sum_{i=2}^{N-2}2 \eta_i E\{ (\theta_i - \theta^* - t_i s_i)^\top (\theta_i - \theta_{i-1})\} + \sum_{i=2}^{N-2}\eta_i^2 E\{\| \theta_i - \theta_{i-1} \|^2 \}] \nonumber \\
& +  t_{N-1}^2 (M + \sigma^2)  + 2 \eta_{N-1} E\{ (\theta_{N-1} - \theta^* - t_{N-1} s_{N-1} )^\top (\theta_{N-1}- \theta_{N-2}) \} \nonumber \\
& + \eta_{N-1}^2 E \{ \| \theta_{N-1}- \theta_{N-2}\|^2\} \nonumber \\
& \leq e^{ -(m \sum_{j=1}^{N-1} t_j + \frac{m^2}{2} \sum_{j=1}^{N-1}t_j^2)}[E\{ \|\theta_2 - \theta^*\|^2\}+\sum_{i=2}^{N-2} t_i^2 (M + \sigma)^2 \nonumber \\
& +\sum_{i=2}^{N-2}2 \eta_i E\{ (\theta_i - \theta^* - t_i s_i)^\top (\theta_i - \theta_{i-1})\} + \sum_{i=2}^{N-2}\eta_i^2 E\{\| \theta_i - \theta_{i-1} \|^2 \}] \nonumber \\
& +  t_{N-1}^2 (M + \sigma^2)  + 2 \eta_{N-1} E\{ (\theta_{N-1} - \theta^* - t_{N-1} s_{N-1} )^\top (\theta_{N-1}- \theta_{N-2}) \} \nonumber \\
& + \eta_{N-1}^2 E \{ \| \theta_{N-1}- \theta_{N-2}\|^2\}.
\end{align}
Under Assumptions 1 and 3, every term in sequences $\{E \{ \| \theta_{i}- \theta_{i-1}\|^2\}\}_i$ and $\{E\{ (\theta_{i} - \theta^* - t_{i} s_{i} )^\top (\theta_{i}- \theta_{i-1}) \}\}_i$ are bounded. Because $\sum_i t_i \rightarrow \infty$, $\sum_i t_i^2 < \infty$, the first, second, and fifth terms in Eq. (\ref{sto:expand}) approach zero as $N$ goes to infinity. Because $\{t_i\}$ and $\{\eta_i < 1\}$ are non-increasing diminishing-to-zero sequences, the sixth and seventh terms in Eq. (\ref{sto:expand}) approach zero, as $N \rightarrow \infty$. Finally, if 
$e^{ -(m \sum_{j=1}^{N-1} t_j + \frac{m^2}{2} \sum_{j=1}^{N-1}t_j^2)} \sum_i \eta_i  \rightarrow 0$, then $ e^{-m \sum_i t_i}\sum_i \eta_i^2 \rightarrow 0$ (the fourth term). Thus, we obtain $E\{ \| \theta_{N} - \theta^*\|^2\} \rightarrow 0$ as $N \rightarrow \infty$. Eq. (\ref{diminishingt}) is obtained because the dominating terms, as $N$ becomes large, are the first and third terms of Eq. (\ref{sto:expand}), and $c_0 $ in Eq. (\ref{diminishingt}) is a number greater than $\max\{\max_i 2 \{E\{ (\theta_i - \theta^* - t_i s_i)^\top (\theta_i - \theta_{i-1})\}\},E\{ \|\theta_2 - \theta^*\|^2\}\}$. 

\noindent (i) $\sum_i \eta_i < \infty$: It is clear that the convergence rate is the same as that of SG (Lemma \ref{SGcon}).

\noindent (ii) $\eta_i = \frac{1}{j+1}$: We consider individual convergence rate of $e^{-(\sum_i^{N} t_i + \frac{1}{2} \sum_j^N t_j^2) }$ and $\sum_i^N \eta_i$. As analyzed in Lemma \ref{SGcon}, the order of $e^{-(\sum_i^{N} t_i + \frac{1}{2} \sum_j^N t_j^2) }$ is ${\cal O}(\frac{1}{N+1})$. Also, $\log N$ is the upper bound of $\sum_i^N \eta_i$ with $\eta_i =  \frac{1}{j+1}$, due to 
\begin{align*} 
\sum_{j}^{N+1} \frac{1}{j+1} \leq \int_{1}^{N+1} \frac{1}{x} \;dx = \log (N+1).
\end{align*}
Therefore, the order is ${\cal O}(\frac{\log (N+1)}{N+1})$. \\

\noindent (iii) $\eta_i = \frac{1}{(j+1)^{\beta}}$ with $0 < \beta < 1$: Individual convergence rate of $e^{-(\sum_i^{N} t_i + \frac{1}{2} \sum_j^N t_j^2) }$ and $\sum_i^N \eta_i$ is considered. The order of $e^{-(\sum_i^{N} t_i + \frac{1}{2} \sum_j^N t_j^2) }$ is ${\cal O}(\frac{1}{N+1})$ (Lemma \ref{SGcon}), while $\frac{N^{1-\beta}}{1-\beta}$ is an upper bound of $\sum_i \frac{1}{(1+ j)^\beta}$ for all $N$, due to
\begin{align} \label{slowhar}
\sum_{i=1}^{N+1} \frac{1}{(1+ j)^\beta} \leq \int_1^{N+1} \frac{1}{x^{\beta}} \; dx \leq  \frac{1}{1-\beta} (N+1)^{1-\beta}.
\end{align}
Combining the two individual rates, we obtain order ${\cal O}(\frac{1}{(1-\beta) (N+1)^{\beta}})$.

\qed

\section{Proof of Eq. (\ref{constantM}) } \label{appendixC}

Eq. (\ref{SGkey1}) is copied below for convenience: 
\begin{align} \label{sto:keyeq1}
E\{\|\theta_{N} -\theta^*\|^2\} & \le (M + \sigma^2) \sum_{i=1}^{N-2} [t_i^2 \prod_{j=i+1}^{N-1}(1-t_j m)]  + (M + \sigma^2) t_{N-1}^2  \nonumber \\
& + \prod_{j=1}^{N-1} (1-t_j m) E\{\|\theta_{1} -\theta^*\|^2\}.
\end{align}
We let 
\begin{align} \label{cons}
0 < 1- a m = \alpha < 1.
\end{align}
The first and second terms on the right-hand side of Eq. (\ref{sto:keyeq1}) can be bounded for any $N$:
\begin{align} \label{const1}
(M + \sigma^2) ( \sum_{i=1}^{N-2} [a^2 \prod_{j=i+1}^{N-1}(1-a m)]  + a^2)
&= (M + \sigma^2)a^2( \sum_{i=1}^{N-2} \alpha^{N-i-1} + 1) \nonumber \\
& \leq (M+ \sigma^2) a^2 \frac{1}{1-\alpha}.
\end{align}
We substitute this bound back into Eq. (\ref{sto:keyeq1}) and use Eq. (\ref{cons}) to obtain
\begin{align} 
E\{\|\theta_{N} -\theta^*\|^2\} 
& \le  (M + \sigma^2)a^2 (\frac{1}{1-\alpha}) 
+ \alpha^{N-1} E\{\|\theta_1 -\theta^*\|^2\}. \label{esti}
\end{align}
Because $\alpha < 1$ and the first term on the right-hand of Eq. (\ref{esti}) is a constant, there exists a finite but sufficiently large $N_a$ to make  
\begin{align} \label{N0}
\alpha^{N_a-1} E\{\|\theta_{1} -\theta^*\|^2\} \leq  (M + \sigma^2)a^2 (\frac{1}{1-\alpha}).
\end{align}
Hence, for $N \geq N_a$, Eq. (\ref{esti}) is
\begin{align*}
E\{\|\theta_{N} -\theta^*\|^2\}  \leq  (M + \sigma^2)a^2 (\frac{2}{1-\alpha}) \leq \frac{2(M + \sigma^2)}{m} a.
\end{align*}

\section{Proof of Theorem \ref{piecewisemomlem}} \label{piecewisemom}

Taking the sum of Eq. (\ref{expectM}) with $i=0$ to $j$, we obtain
\begin{align*}
E\{\|\theta_{j+1} -\theta^*\|^2\}  \leq E\{\|\theta_0 -\theta^*\|^2\} - m \sum_{i=0}^j t_i E\{\|\theta_i -\theta^*\|^2\} + (M + \sigma^2) \sum_{i=0}^j t_i^2  \\
+ 2 \sum_{i=0}^j \eta_i E\{(\theta_i -  \theta^* - t_i s_i )^\top (\theta_i- \theta_{i-1}) \}+ \sum_{i=0}^j \eta_i^2 E\{\|\theta_i - \theta_{i-1}\|^2\}.
\end{align*}
Rearranging the above and multiplying the result with $\frac{1}{\sum_{i=0}^j t_i}$ yields
\begin{align*}
 \frac{m\sum_{i=0}^j t_i E\{\|\theta_i -\theta^*\|^2\}}{\sum_{i=0}^j t_i}\leq  
(M + \sigma^2) \frac{\sum_{i=0}^j t_i^2}{\sum_{i=0}^j t_i}-  \frac{(E\{\|\theta_{j+1} -\theta^*\|^2\} - E\{\|\theta_0 -\theta^*\|^2\})}{\sum_{i=0}^j t_i} \\
+ \frac{1}{\sum_{i=0}^j t_i}[ 2\sum_{i=0}^j \eta_i E\{(\theta_i -  \theta^* - t_i s_i )^\top (\theta_i- \theta_{i-1}) \}+ \sum_{i=0}^j \eta_i^2 E\{\|\theta_i - \theta_{i-1}\|^2\}]
\end{align*}
If we let $\nu_i = \frac{t_i}{\sum_{i=0}^j t_i}$ and use Jessen's inequality for convex functions, we obtain the following:
\begin{align*}
\sum_{i=0}^j  \nu_i E\{\|\theta_i -\theta^*\|^2\} \geq 
E\{\| \sum_{i=0}^j\nu_i \theta_i -\theta^*\|^2\}.
\end{align*}
We thus have
\begin{align*}
m E\{\| \sum_{i=0}^j \nu_i \theta_i -\theta^*\|^2\} \leq 
(M + \sigma^2) \frac{\sum_{i=0}^j t_i^2}{\sum_{i=0}^j t_i}-  \frac{(E\{\|\theta_{j+1} -\theta^*\|^2\} - E\{\|\theta_0 -\theta^*\|^2\})}{\sum_{i=0}^j t_i} \\
+  \frac{1}{\sum_{i=0}^j t_i} [2\sum_{i=0}^j \eta_i E\{(\theta_i -  \theta^* - t_i s_i )^\top (\theta_i- \theta_{i-1}) \}+ \sum_{i=0}^j \eta_i^2 E\{\|\theta_i - \theta_{i-1}\|^2\}]
\end{align*}
Now, let $t_i = a$. We obtain $\nu_i = \frac{1}{j+1}$ for $i \leq j$ and
\begin{align} \label{SGMconstant}
m E\{\|\frac{1}{j+1} \sum_{i=0}^j \theta_i -\theta^*\|^2\}  \leq 
(M + \sigma^2) a -  \frac{(E\{\|\theta_{j+1} -\theta^*\|^2\} - E\{\|\theta_0 -\theta^*\|^2\})}{(j+1)a} \nonumber \\
 +  \frac{1}{ (j+1)a} [2\sum_{i=0}^j \eta_i E\{(\theta_i -  \theta^* - t_i s_i )^\top (\theta_i- \theta_{i-1}) \}+ \sum_{i=0}^j \eta_i^2 E\{\|\theta_i - \theta_{i-1}\|^2\}] 
\end{align}
Under Assumptions 1-3, $E\{\|\theta_i - \theta_{i-1}\|^2\}$ and $\|E\{(\theta_i -  \theta^* - t_i s_i )^\top (\theta_i- \theta_{i-1}) \}\|$ are bounded for all $i$. Because $j a \rightarrow \infty$ and $\eta_i < 1$, from $\lim_{j \rightarrow \infty} \frac{\sum_{i=0}^j \eta_i}{(j+1)a} \rightarrow 0$, we can deduce 
$\lim_{j \rightarrow \infty} \frac{\sum_{i=0}^j \eta_i^2}{(j+1)a} \rightarrow 0$ and that there exists a finite number $N_a$ such that for $N \geq N_a$, 
\begin{align*}
a (M + \sigma^2) & \geq 
\frac{1}{(j+1)a}[
E\{\|\theta_0 -\theta^*\|^2\} \\
& + [2\sum_{i=0}^j \eta_i E\{(\theta_i -  \theta^* - t_i s_i )^\top (\theta_i- \theta_{i-1}) \}+ \sum_{i=0}^j \eta_i^2 E\{\|\theta_i - \theta_{i-1}\|^2\}].
\end{align*}
Substituting this into Eq. (\ref{SGMconstant}) yields 
\begin{align*}
E\{\|\frac{1}{N+1} \sum_{i=0}^N \theta_i -\theta^*\|^2\}  \leq  \frac{2(M + \sigma^2)}{m} a
\end{align*}
for $N \geq N_a$.

\end{document}